# A Self-Supervised Miniature One-Shot Texture Segmentation (MOSTS) Model for Real-Time Robot Navigation and Embedded Applications

Yu Chen, Chirag Rastogi, Zheyu Zhou, and William R. Norris, Member, IEEE

*Abstract*— Determining the drivable area, or free space segmentation, is critical for mobile robots to navigate indoor environments safely. However, the lack of coherent markings and structures (e.g., lanes, curbs, etc.) in indoor spaces places the burden of traversability estimation heavily on the mobile robot. This paper explores the use of a self-supervised one-shot texture segmentation framework and an RGB-D camera to achieve robust drivable area segmentation. With a fast inference speed and compact size, the developed model, MOSTS is ideal for real-time robot navigation and various embedded applications. A benchmark study was conducted to compare MOSTS's performance with existing one-shot texture segmentation models to evaluate its performance. Additionally, a validation dataset was built to assess MOSTS's ability to perform texture segmentation in the wild, where it effectively identified small low-lying objects that were previously undetectable by depth measurements. Further, the study also compared MOSTS's performance with two State-Of-The-Art (SOTA) indoor semantic segmentation models, both quantitatively and qualitatively. The results showed that MOSTS offers comparable accuracy with up to eight times faster inference speed in indoor drivable area segmentation.

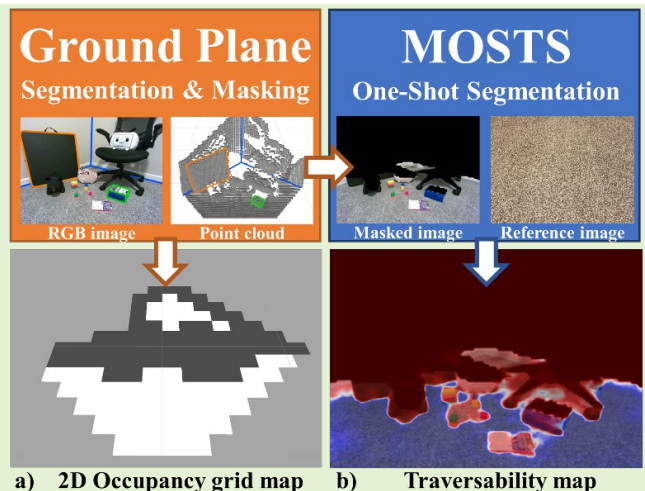

Fig. 1. The outputs of the framework: a) a 2D grid map that captures large obstacles b) a traversability map (**blue** indicates drivable area)

## I. Introduction

Effective robot perception is a critical component of successful mobile robot navigation, particularly for indoor robots operating in cluttered and confined spaces designed for human use. These robots face unique challenges, such as limited capacity for carrying batteries and computers, complex indoor layouts that impede sensor performance, and the risk of tipping or jamming due to small obstacles given their low ground clearances.

This research lab is developing a personal mobility device named MiaPURE (Modular, Interactive, and Adaptive Personalized Unique Rolling Experience, Fig.2). MiaPURE is an assistive rideable ballbot system, characterized by low ground clearance and driven by a single spherical wheel. Besides facing the challenges common to traditional ground robots, MiaPURE also contends with under-actuated dynamics, which induces tilting motions during acceleration [1]. The commonly held assumption that the transformation between the ground plane and the robot's sensor coordinates is (entirely or partly) constant and known [2], [3] does not apply to dynamically balanced devices like MiaPURE, as these robots are not constrained to strict planar motion.

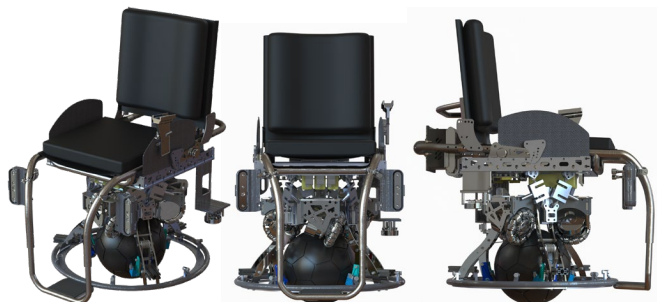

Fig. 2. The assistive rideable ballbot system, MiaPURE

Inspired by the need to safeguard MiaPURE, this paper proposes an efficient indoor Drivable Area Segmentation (DAS) framework that is robust to tilting motion and capable of detecting small unknown obstacles in real-time on UpXtreme-

This work was supported by NSF NRI-2.0 #2024905. (Corresponding author: Yu Chen.)

Yu Chen and Zheyu Zhou are with the Department of Mechanical Science & Engineering, University of Illinois Urbana-Champaign, Urbana, IL 61801 USA (e-mail: yuc6@illinois.edu, zheyuz3@illinois.edu).

Chirag Rastogi is with the Department of Computer Engineering, University of Illinois Urbana-Champaign, Urbana, IL 61801 USA (e-mail: chiragr2@illinois.edu).

William R. Norris is with the Department of Industrial & Enterprise Systems Engineering, University of Illinois Urbana-Champaign, Urbana, IL 61801 USA (e-mail: wrnorris@illinois.edu).

i7, a low-power Single Board Computer (SBC). The proposed method approaches the problem of DAS in two key steps. First, a real-time and motion-robust ground plane segmentation node segments the point cloud collected by an RGB-D camera and makes an initial estimation of the drivable area. Second, a one-shot segmentation model refines the drivable area by filtering out small obstacles and anomalies in the RGB image. The outputs of the proposed method are a simplistic 2D obstacle map that encodes all the large obstacles (e.g., walls, furniture, pedestrians, etc.) and a traversability map that indicates the surrounding regions that are free of small obstacles or anomalies. With a reference texture of the traversable surface, this approach seeks to identify all potential obstacles threatening the robot system, including small, unknown objects that may not be detectable using point clouds or depth images.

Recently, there has been a growing trend of training large Vision Transformer (ViT) models, such as the Segment Anything Model (SAM) [4], on unprecedentedly large datasets to achieve unparalleled accuracy and generalizability. However, these models often require extensive computational and data resources, making them impractical for real-time deployment on low-power and embedded systems. Additionally, the resource requirements for training these models are often beyond the reach of many researchers and individuals. In contrast to this approach, this paper focuses on developing a lightweight and versatile model that can be trained and deployed with significantly fewer resources. The training of the proposed method did not rely on any dense (pixel-wise annotations) manually labeled dataset as it followed a self-supervised methodology. The short inference time enabled the proposed method's deployment on an SBC (93.13 FPS for the neural network model alone and about 30 FPS for the entire framework due to the limited sampling rate of the camera). By using a one-shot segmentation formulation, this approach allows the user to specify traversable regions since not all floor areas are suitable for indoor robots. Furthermore, one can re-configure the proposed framework to adapt to a different working environment or robot mission by changing the reference image during runtime (Fig.3).

The key contributions of this paper include the following:
1. A Miniature One-Shot Texture Segmentation (MOSTS) model optimized for an embedded, real-time application.
2. A novel Perlin-Noise-based [5] collage generation technique for training robust texture segmentation models.
3. An Indoor Small Objects Dataset (ISOD) that contained diverse and densely labeled images for validating the performance of the proposed method in the wild.

The code and dataset used in this study are publicly available at: https://github.com/mszuyx/MOSTS

The rest of this paper is organized as follows: Section II. is a brief literature review of related studies. Section III. offers an overview of the proposed framework. Section IV. and Section V. provide the implementation details of the ground plane segmentation algorithm and the one-shot DAS model. Section VI. explains the experimental setup. Section VII. presents and discusses the validation results, followed by conclusions and suggestions for future work in Section VIII.

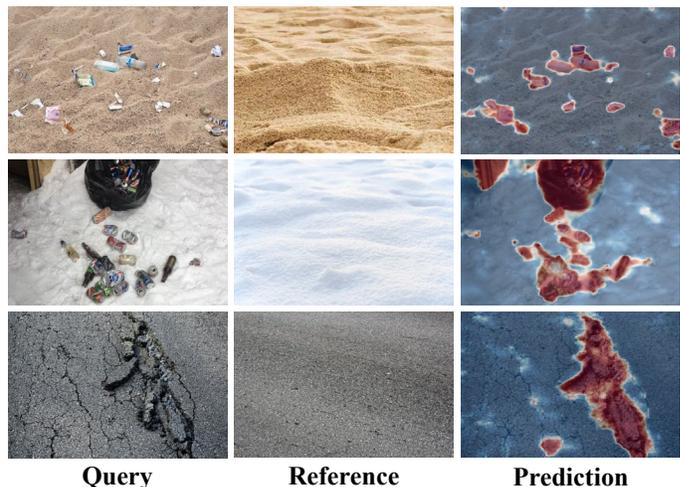

**Query**      **Reference**      **Prediction**

Fig. 3. One-shot texture segmentation in the wild using MOSTS. Query and reference images were collected from the internet.

## II. RELATED WORK

### A. Indoor Vision-based Drivable Area Segmentation

In some earlier attempts, [6], [7] used vision-based methods that segmented the obstacle-free floor from monocular images based on the assumption that the floor has a uniform texture. More recently, [8]–[10] approached this problem using supervised Deep Neural Networks (DNNs) and achieved promising results. However, training these DNNs required a large amount of expensive manually labeled data.

Furthermore, the authors of [11] pointed out that the difficulty of indoor DAS went beyond the lack of coherent vision marks. The study in [11] found that existing homography-based floor segmentation methods often failed to capture low-lying small obstacles (typically $\leq 3\ cm$ height) in RGB images. The authors further stated that it would also be difficult for depth sensors such as laser range finders and stereo cameras to detect low-lying obstacles based on noisy depth values.

### B. One-Shot Texture Segmentation

Textures are effective visual priors humans use to perform daily low-level visual inference and scene understanding [12], [13]. The use of texture representations is also very common in applications like terrain recognition [14] and surface defect (or anomaly) detection [15]. While the studies on texture classification and segmentation are extensive, one-shot texture segmentation (or retrieve) is a relatively recent field of research. The studies in [12], [13] leveraged a Voronoi-based collage generation technique to automatically generate pixel-wise annotations to train their one-shot texture segmentation models. The trained models can identify the region containing the targeted texture when given a reference image.

While [13] was intended for texture retrieval in fashion and e-commerce, this study explored using one-shot texture segmentation for indoor real-time robot navigation. A benchmark study demonstrated that the developed model, MOSTS, outperformed the SOTA models in accuracy and speed. Additionally, this study proposes a novel Perlin-Noise-based collage generation method that significantly improved the robustness of MOSTS.

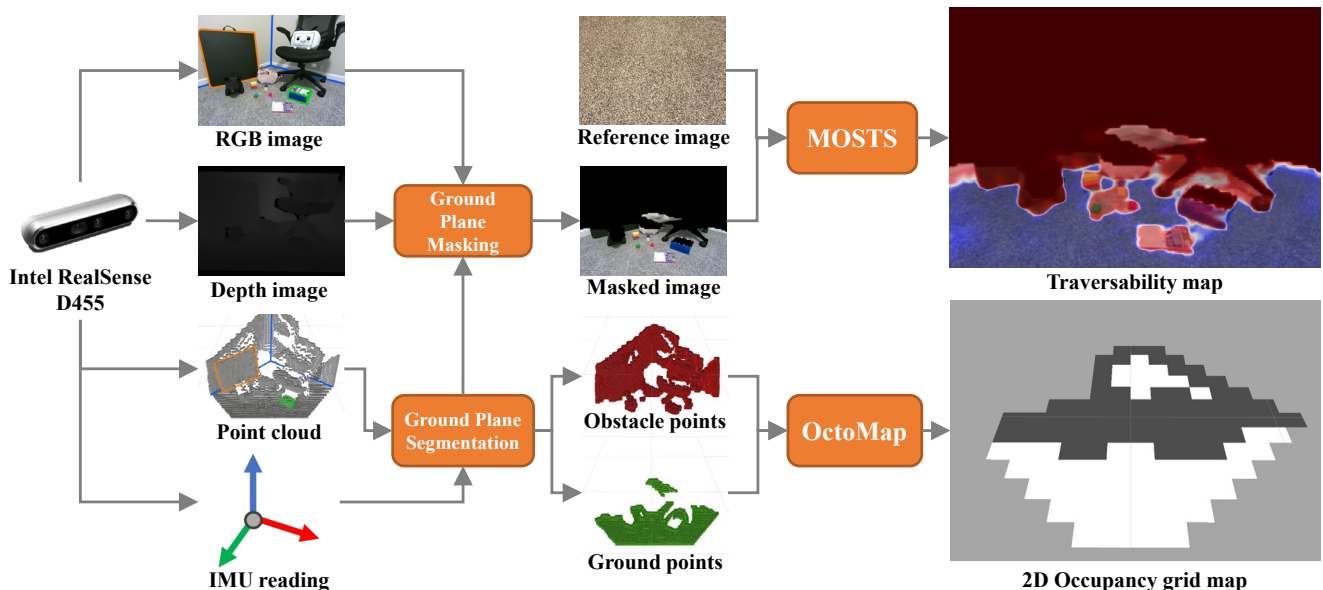

Fig. 4. The overall framework of the proposed method.

### C. Efficient RGB-D Semantic Segmentation

RGB-D semantic segmentation models are powerful and versatile tools in computer vision. However, most of them are not designed for use in robots or embedded systems. This limitation has led to the development of efficient RGB-D semantic segmentation models such as RedNet [16] and ESANet [17]. These models aim to balance adequate accuracy with optimized computation requirements. As suggested in [17], these models can enrich the robot's visual perception and facilitate subsequent processing steps, like DAS, based on semantic information. The method proposed in this study performs DAS based on comparing texture differences rather than relying on semantic or contextual information from the scenes. This approach makes the method more generalizable to unknown environments and objects. Depending on their training dataset distribution, traditional semantic segmentation models might struggle to identify small unknown objects. However, this study demonstrated that the proposed method can accurately identify these objects while offering significantly faster inference times.

In quantifying the efficacy of the proposed method, this study compared its results with [16] and [17] as they are considered good performance indicators (baselines) for indoor DAS.

### III. OVERVIEW

Fig.4 provides the overall framework of the proposed method and visualization for each data stream. The RGB-D camera (Intel RealSense D455) offers four types of sensor data: the RGB image, the depth image and the point cloud that represents the 3D structures of the captured scene, and the Inertial Measurement Unit (IMU) reading that describes the orientation of the camera. The proposed framework feeds the point cloud and IMU data to a ground plane segmentation node that separates the raw point cloud into a ground plane group and a non-ground group. The segmented point clouds are used to construct a 3D Octomap (i.e., a grid map made by 3D voxels) [18] that describes the 3D scene. Then, the 3D voxels are projected to the ground to generate a more simplistic 2D occupancy grid for subsequent high-level control tasks like path planning and navigation. The ground plane model and the depth image are used to mask the raw RGB image, such that most of the obstacle pixels are masked out (painted black). This pre-processing step prepares the query image for MOSTS, which further segments the ground regions into drivable areas and anomalies based on a target terrain surface given by a reference image. The outcome of MOSTS can be re-projected to the robot workspace coordinate to form a traversability map.

### IV. REAL-TIME MOTION-ROBUST OBSTACLE GRID MAP

### A. Problem Statement

The ground plane segmentation of 3D point clouds is a common and essential step to filter the points with less interest [19], [20]. In the context of this paper, ground plane segmentation was used as a pre-processing step to obtain a preliminary estimation of the drivable areas. As the proposed framework is expected to operate primarily indoors, this study made the approximation that the geometry of an indoor ground plane can be roughly modeled using a point-normal equation:

$$ax + by + cz + d = 0 \qquad (1)$$

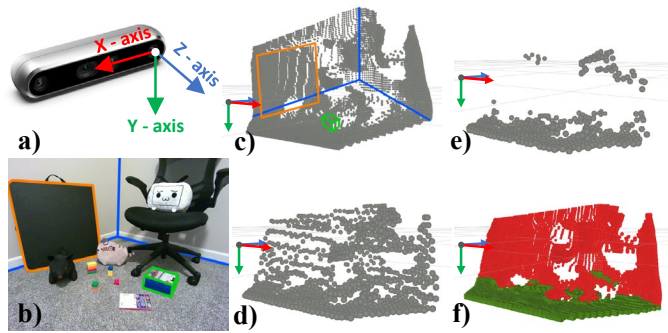

Fig. 5. a) The point cloud frame definition; b) the RGB image of the scene; c) the raw input point cloud; d) the point cloud after IMU alignment and the voxel grid filter; e) the point cloud after the radius outlier removal filter; f) the final output, **green** indicates ground points, and **red** indicates obstacle/non-ground points. The corner of the shown scene and some objects are highlighted in b) and c) for easier visual understanding. The coordinate frame in c)-f) indicates the camera position and orientation.

where $[a, b, c]^T$ are constant values that form the normal vector $\vec{n}$ of the ground plane, and $d$ can be seen as the orthogonal offset distance between the ground plane and the point cloud's frame. The objective of the ground plane segmentation node is to estimate the values of the normal vector $\vec{n}$ and offset distance $d$ such that they can be used to determine whether a given point belongs to the ground plane.

### B. Filtering

Similar to [19]–[21], the proposed ground plane segmentation algorithm used a RANdom SAmple Consensus (RANSAC) based method. For this method to succeed, the most crucial part is to select a reasonably good initial guess (candidate points) for the targeted plane. This can reduce the need for more RANSAC iterations, and the algorithm is more likely to converge to the true plane. In reliably obtaining these candidate points, the proposed method applied a series of filters and operations to the input point cloud (Fig.5 c-f).

First, the camera's IMU reading (pitch and roll angles) was used to align the point cloud orientation with respect to the gravity vector to cancel the robot's tilting motion.

Second, a voxel grid filter was used to downsample the point cloud (Fig.5d). The grid size was empirically defined to be *0.03m × 0.2m × 0.03m* (X-Y-Z convention) such that the output point cloud had a lower resolution on the vertical axis (i.e., y-axis, the coordinate frame is defined in Fig.5a).

Finally, a radius outlier removal filter was used. This filter used a Kd-Tree radius search to determine the number of neighbors $k$ that lay within a radius of a query point. If $k$ was less than a threshold, this query point was removed. This filter helped reduce artifact points created by reflective surfaces. Also, as the filtered point cloud had a lower resolution on the vertical axis (y-axis), points in vertical structures (e.g., walls, large furniture) were sparse and had fewer neighboring points. Hence, they were more likely to be filtered out. Most remaining points likely belonged to large planes orthogonal to the gravity vector (Fig.5e) and made good RANSAC candidates.

### C. Ground Plane Segmentation

Once the candidate points were determined, they were sorted based on their heights, and the lowest 50% of points were sampled to form the initial seed for the RANSAC algorithm. The RANSAC algorithm then iteratively estimated the ground plane model $\vec{n}$ and $d$ using the candidate points. Ideally, $\vec{n}$ can be derived directly from the IMU readings. However, there is no guarantee that the floor plane is perfectly orthogonal to the gravity vector, and the IMU readings are also susceptible to strong vibrations during operation. Therefore, a Principal Component Analysis (PCA) was used to refine the normal vector $\vec{n}$ by finding the singular vector corresponding to the smallest singular value. As the candidates were pre-filtered using the IMU, RANSAC was less likely to converge onto false planes (e.g., walls, large furniture) than the PCA method alone.

At the final iteration, the algorithm used the estimated $\vec{n}$ and $d$ to infer the unfiltered point cloud and segment it into two groups: the obstacle group (red) and the ground plane group (green) (Fig.5f). In general, the entire segmentation algorithm required $6\sim13ms$ computation time on the SBC. Typically, this setup can recognize floor objects higher than $5cm$.

### D. Ground Plane Masking

The estimated model in (1) was used to mask the RGB image and form the "masked image" in Fig.4, which served as one of MOSTS's inputs. MOSTS is designed to segment the regions in the query image that share a similar texture class to the reference image and does so without any scene understanding. When the surrounding walls or furniture have a similar texture as the ground/floor, MOSTS might mistake those regions as drivable areas. However, using the masked ground image can prevent these edge cases.

## V. TEXTURE ORIENTED, SELF-SUPERVISED ONE-SHOT DRIVABLE AREA SEGMENTATION

### A. Problem Statement

Similar to other one-shot segmentation models, this model takes two inputs, i.e., the query image Q, which contains regions of different textures, and the reference image R, which embodies the target texture class. The output of MOSTS is a probability map that indicates regions of the query image that correspond to the target texture class. To train such a model, a triple $T^i$ of training samples was required:

$$T^i = (Q, R^i, G^i) \qquad (2)$$

Where Q is the query image that contains at less one texture class, $G^i$ is the pixel-wise ground truth label corresponding to the $i^{th}$ class texture in Q, and $R^i$ is the reference image of the $i^{th}$ class. The neural network model $f(*)$ can be presented in the following form:

$$P^i = f(Q, R^i, \theta) \qquad (3)$$

Where $P^i$ is the predicted segmentation result, and $\theta$ is the set of learnable parameters in function $f(*)$. The goal of the training process is to learn the optimized $\theta^*$ given the training triple $T^i$ such that the loss function is minimized:

$$\theta^* = \underset{\theta}{\operatorname{argmin}} \, Loss(P^i, G^i) \qquad (4)$$

This study used the Combo Loss [21] as the loss function. It is essentially a weighted sum of the B-BCE and Dice Loss. The combination of these two loss functions is smooth and able to handle class imbalance [22].

### B. Perlin-Noise-based collage generation & Model Training

[12] and [13] used a benchmark texture database called the Describable Textures Dataset (DTD) [23] to train their models. DTD is a dataset that contains 5640 images organized into 47 texture categories inspired by human perception. A random Voronoi-based collage generation technique was used in both

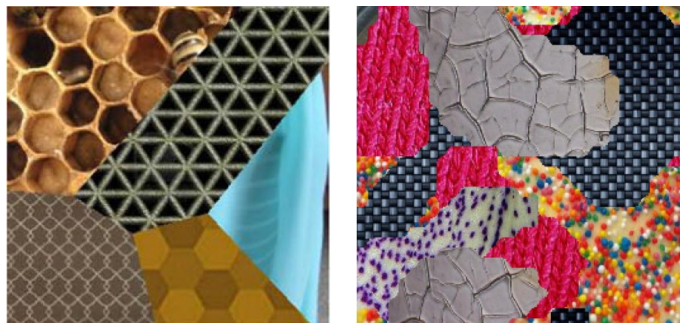

a) Voronoi-based collage    b) Perlin-Noise-based collage

Fig. 6. A comparison between the generation techniques.

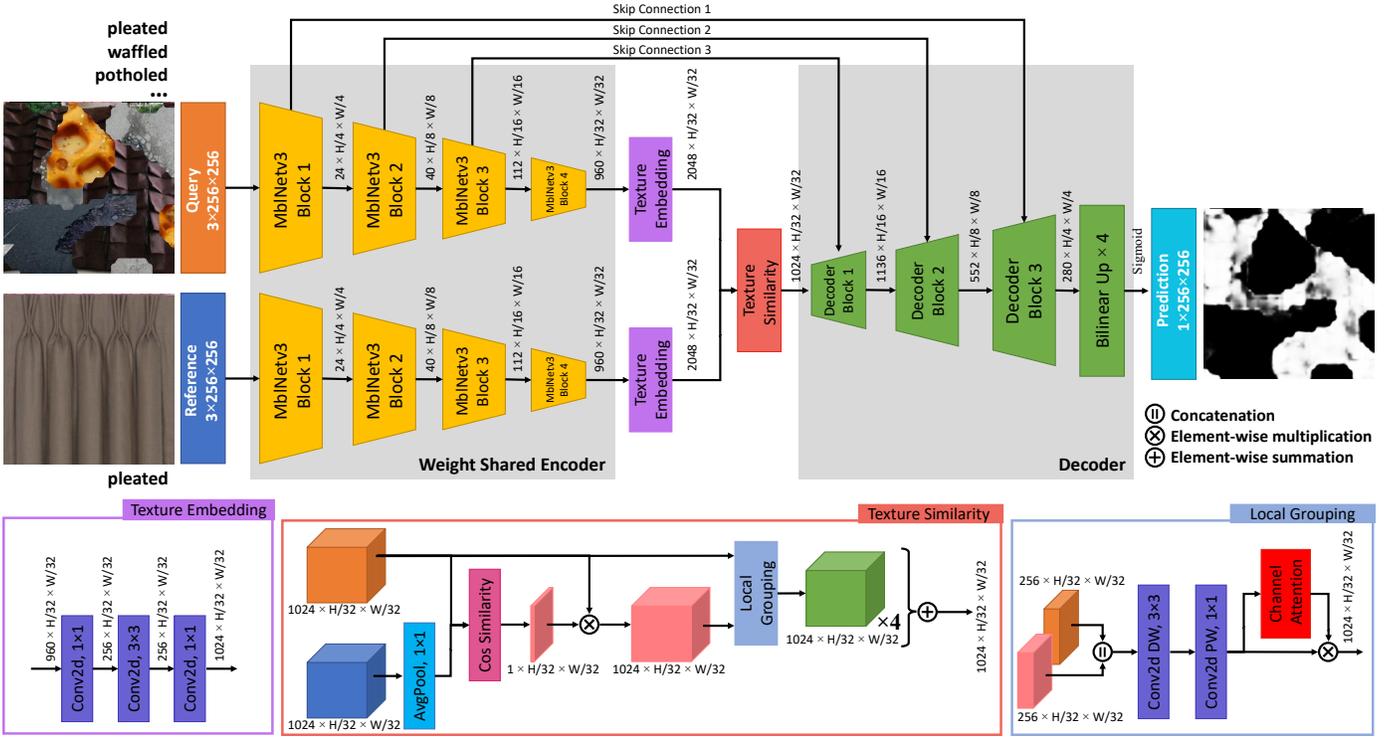

Fig. 8. The architecture of MOSTS.

[12] and [13] to form synthetic query images (Fig.6.a). However, query images generated in this manner generally have three biases: 1) there is no discontinuous region of the same textures; 2) there are only straight edges separating different textures; 3) there is no small region in the collage, and regional areas are relatively balanced. As a result, the model trained using these synthetic query images likely lacks the ability to deal with arbitrary class boundaries in the real world. Therefore, this study designed a novel Perlin-Noise-based collage generator to synthesize pseudo-random query images (Fig.6.b).

Perlin noise is a procedural generation technique widely used in the game development and visual effect industries to generate pseudo-random textures or terrain maps. The smoothness and the amount of detail in the generated shapes can be manipulated by the grid size and the number of Octaves used in the Perlin noise function. This allows the user to have some degree of control over the random generation process. This generator creates a pseudo-random binary map that masks each texture candidate in the query image. Then, all masked candidates are concatenated to form the final synthetic query image (Fig.7).

During training, all images were resized to $256 \times 256$, and up to five texture candidates will be randomly sampled to form the query image. A reference image $R^i$ that corresponds to the $i^{th}$ class (the target class) in Q was randomly chosen. The pseudo-random mask of the $i^{th}$ class is used as $G^i$. Together, Q, $R^i$, and $G^i$ form the triple $T^i$ in (2).

All triples were randomly generated at each training or evaluation step. All random processes were tightly controlled by random seeds and random generators to ensure reproducibility. Similar to [13], among the 47 texture classes in DTD, 42 of them were used to form the training set, and the remaining five classes were used as an evaluation set.

### C. Model Building & Ablation Study

*Encoder:* MOSTS employs a typical dual-branch encoder structure (Siamese encoder) found in many one-shot learning models (Fig.8). Since the encoder's throughput significantly contributes to the model inference time, the selection of an efficient backbone model for the encoder is vital for achieving real-time performance. This study conducted a benchmark test to identify the most suitable backbone model using seven popular models commonly utilized for building efficient segmentation models. The selected models include ResNet-based models[24] (i.e., ResNet-18, 34, and 50), EfficientNet [25], EfficientNet-v2 [26], MobileNet-v2 [27], and MobileNet-v3 [28]. All models were pre-trained on the ImageNet [29] dataset and converted to Float16 Intermediate Representation using the model optimizer from Intel OpenVINO (ver 2022.1). As the SBC used in this study is powered mostly by Intel hardware (Intel i7-8665UE CPU and Intel UHD Graphics 610 GPU), OpenVINO IR can provide optimal performance. Fig.9 presents a bubble plot of the benchmark results for these models, with the bubble sizes indicating the relative sizes of each model. As observed from Fig.9, the chosen model, MobileNet-v3, demonstrated an inference speed more than twice as fast as the most accurate model (EfficientNet-v2), while showing only a 3% reduction in accuracy.

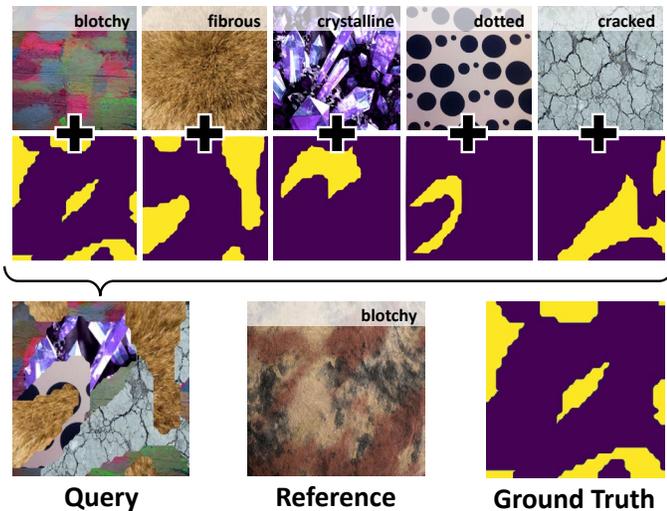

Fig. 7. The Perlin-Noise-based collage generation process. The **yellow** regions have pixel values of 1, whereas the **purple** regions are 0.

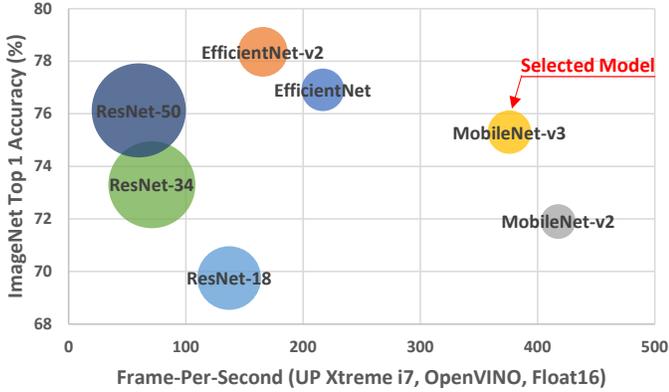

Fig. 9. The benchmark result of the backbone models on ImageNet with input size 224×224.

*Texture Embedding:* Given that the backbone encoder model was pre-trained on ImageNet for classification, its feature embedding might not provide optimal performance for texture segmentation tasks. Meanwhile, fine-tuning the backbone model could potentially jeopardize model robustness and generalizability, considering that the training dataset, DTD, is considerably less diverse than ImageNet. To overcome these limitations, a texture embedding block was introduced to refine the initial features into more appropriate texture feature embeddings. As illustrated in Fig.8, this block incorporates a $3 \times 3$ convolution layer flanked by two $1 \times 1$ convolution layers. Each convolution operation is succeeded by Batch Normalization and a ReLU activation function. This convolution block adopts a bottleneck structure, effectively compressing the channel dimension from its original size of 960 to 256. This design encourages the model to learn to extract vital information from preliminary features efficiently.

*Texture Similarity:* Inspired by [30], the texture similarity block use cosine similarity as an attention mechanism (Fig.8) to emphasize the features of the query texture embedding if they show strong similarities with the reference texture embedding in the spatial dimensions. First, an average pooling operation compresses the spatial dimensions of the reference texture embedding into a $1024 \times 1 \times 1$ tensor. Given that a visual texture essentially consists of a set of repeated primitive texels arranged on the image plane, its fundamental pattern is inherently spatially invariant. If the reference image effectively captures the essence of the target texture, the compressed tensor can be viewed as an encoded texel. This tensor is then used to compute the cosine similarity $s_{x,y}$ with the query texture embedding at each spatial position, enabling the identification and emphasis of similar texture features:

$$s_{x,y} = \frac{r \cdot F^q_{x,y}}{\|r\|_2 \cdot \|F^q_{x,y}\|_2} \quad (5)$$

The compressed reference tensor is denoted as $r \in R^{C \times 1 \times 1}$ and the feature embedding of the query image is $F^q \in R^{C \times w' \times h'}$. The computed similarity map $s \in R^{1 \times w' \times h'}$ encodes the similarity values $[-1,1]$ at each spatial position $(x, y)$. The model masked the query texture embedding using the similarity map via element-wise multiplication.

*Local Grouping:* The local grouping block was initially used in [13] to aggregate features learned by different convolution branches. This study used it to aggregate the masked query features with the unmasked features to preserve information prior to the texture similarity block. It also allows gradients to bypass the similarity block to refine the texture embedding directly. By organizing feature embeddings into smaller local groups, the model can keep related feature channels close and significantly decrease the number of convolution filters required. In this module, both masked and unmasked query texture embeddings are divided into four groups. Within each group, the embedding pair are concatenated and passed through a depth-wise (DW) and point-wise (PW) convolution block for aggregation (Fig.8). This specific structure was chosen for its efficiency. To further optimize feature selection, a learnable channel-wise attention module [31] was incorporated. This module lets the model focus on useful features while suppressing the less relevant ones. Finally, the output features from the four local groups were combined using an element-wise summation.

*Decoder:* The decoder up-samples the feature embedding via bilinear interpolation. The up-sampled feature is then concatenated with the residual feature provided by the skip connection. A $1 \times 1$ convolution layer subsequently aggregates these concatenated features. This procedure recurs in the three decoding blocks, followed by a final bilinear interpolation layer. This ensures that the model's output shares the same spatial dimensions as its input (Fig.8).

This paper presents an ablation study to dissect the contribution of each model component. For ease of comparison, this study used the same four-fold validation groups for the DTD dataset defined in [13] (Table I).

TABLE I
VALIDATION GROUP DEFINITION FOR DTD

| Groups | Validation Classes |
|---|---|
| i=0 | 'banded','blotchy','braided','bubbly','bumpy' |
| i=1 | 'chequered','cobwebbed','cracked','crosshatched','crystalline' |
| i=2 | 'dotted','fibrous','flecked','freckled','frilly' |
| i=3 | 'gauzy','grid','grooved','honeycombed','interlaced' |

All the models in the ablation study were trained using the DTD dataset for 80 epochs with a batch size of 16. The training optimizer was SGD with a momentum of 0.9 and a weight decay of 0.0001. The initial learning rate was 0.001, and a step scheduler was used. The training was performed on a pre-built PC with 16G of RAM, an Intel i7-11700 CPU, and an NVIDIA GeForce RTX 3060 GPU (12Gb).

Table II presents the results of the ablation study, justifying the design choices made for each module. A baseline accuracy (measured in mIoU) is established using a naive model, which concatenates the outputs of the Siamese encoder before directing the result to the depth-wise and point-wise convolution block and the decoder. The ablation study reveals that the addition of the Texture Embedding (Tex Emb) module does not independently improve accuracy. However, when combined with the Texture Similarity (Tex Sim) module, there is a noticeable increase in the four-fold average mIoU by 4.2, representing approximately a 9% improvement over the baseline model. The most significant improvement (~14%) is achieved when the Local Grouping module is incorporated alongside the Tex Emb and Tex Sim modules.

TABLE II
ABLATION STUDY

| Model | i=0 | i=1 | i=2 | i=3 | mean |
|---|---|---|---|---|---|
| Baseline | 49.1 | 42.8 | 52.7 | 44.2 | 47.2 |
| +Tex Emb | 48.9 | 42.8 | 48.3 | 43.8 | 46.0 |
| +Tex Sim | 48.4 | 47.3 | 53.4 | 44.1 | 48.3 |
| + Tex Emb & Tex Sim | 53.2 | 46.9 | 55.1 | 50.3 | 51.4 |
| + Tex Emb & Tex Sim & Local Grouping | **55.6** | **49.4** | **59.3** | **50.8** | **53.8** |

### D. Comparison with SOTA one-shot segmentation models

To assess the performance discrepancy between MOSTS and existing models, this study replicated the benchmark test conducted in [13], comparing MOSTS with models [12] and [13] using identical four-fold validation groups as defined in Table I. All models underwent training and testing under conditions that were identical to Section V.C. The results in Table III demonstrate that MOSTS, despite its smaller and simpler structure, outperforms models [12] and [13] across all validation groups, achieving an average mIoU of 53.8. This represents an improvement in average mIoU by approximately

TABLE III
COMPARISON WITH SOTA ONE-SHOT TEXTURE SEGMENTATION MODELS

| Model | i=0 | i=1 | i=2 | i=3 | mean |
|---|---|---|---|---|---|
| OSTS [12] | 27.7 | 36.0 | 29.1 | 29.7 | 30.6 |
| OSTR [13] | 52.6 | 47.3 | 53.8 | 49.6 | 50.8 |
| **MOSTS** | **55.6** | **49.4** | **59.3** | **50.8** | **53.8** |

6% and 76% over models [12] and [13], respectively.

## VI. EXPERIMENTAL SETUP

### A. Validation Dataset

This study aims to explore the performance of MOSTS beyond synthetic data, specifically focusing on the potential application of real-time robot navigation. To quantify the proposed method's performance in real-world scenarios, this study collected and labeled an Indoor Small Objects Dataset (ISOD). ISOD contains 2,000 manually labeled images from 20 diverse sites, each featuring over 30 types of small objects randomly placed amidst the items already present in the scenes. These objects, typically $\leq 3cm$ in height, include LEGO blocks, rags, slippers, gloves, shoes, cables, crayons, chalk, glasses, smartphones (and their cases), fake banana peels, fake pet waste, and piles of toilet paper, among others. These items were chosen because they either threaten the safe operation of indoor mobile robots or create messes if run over. Example images from ISOD can be found in Fig.10. In addition to RGB images, ISOD also includes corresponding depth images and IMU readings. To optimize storage space, the outputs of the ground plane masking node (the "masked image" in Fig.4) were logged as the query images for MOSTS instead of storing raw point clouds. A reference image of each floor type was also recorded using a smartphone. Data collection for ISOD was performed using a handheld device (Fig.11), with the operator videotaping the entire scene while moving around at walking speeds ($\leq 1m/s$). During collection, the device was held at varying heights (ranging from $0.2m$ to $1m$) to introduce height variance. The operator was encouraged to simulate the operations of dynamically balanced robots by incorporating tilting motions. Each data collection session lasted three minutes per scene. One hundred samples were randomly selected and added to the ISOD from the raw data recordings. Subsequently, the data labeling team was instructed to annotate the drivable regions in the RGB images with positive labels (pixel value =1), marking the remaining areas with negative labels (pixel value =0).

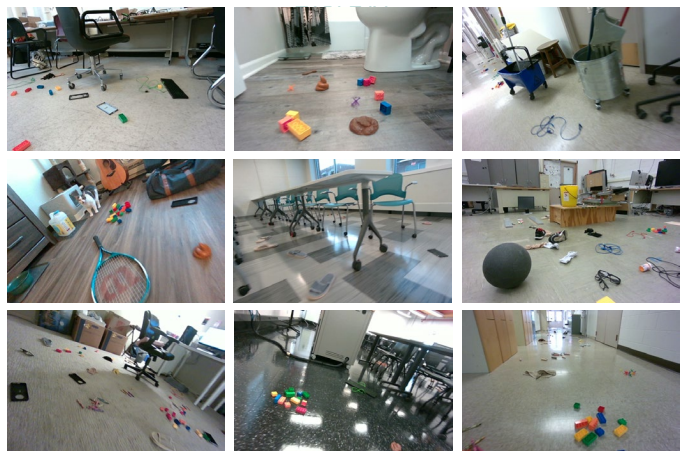

Fig. 10. Example RGB images from ISOD.

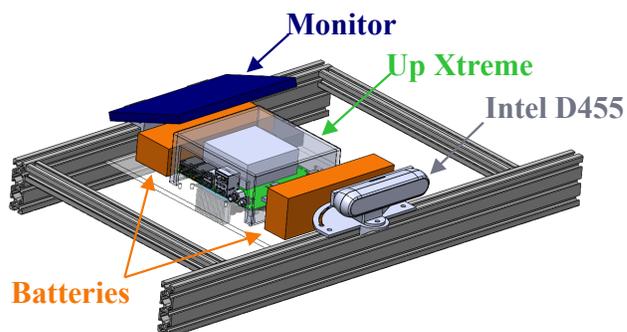

Fig. 11. The handheld device used for data collection.

### B. Preparation for the Benchmark Models

It should be noted that this study used ISOD solely as a validation dataset, while MOSTS was trained utilizing the DTD dataset via the method introduced in Section V.B. Training was conducted for 500 epochs with a batch size of 16, using the ADAM optimizer. Since indoor DAS using one-shot textural segmentation is a relatively recent research area, directly comparing the performance of the proposed method with existing methods proves challenging. Nevertheless, efficient

RGB-D semantic segmentation models such as those detailed in [16] and [17] serve as valuable performance indicators to understand and quantify the framework's efficacy.

As indoor semantic segmentation arguably poses a more complex task than DAS, unmodified versions of models [16] and [17] may pay unnecessary attention to objects unrelated to the DAS task. To ensure a fair comparison, this study retrained models [16] and [17] as DAS models using modified data from SUNRGB-D [32], a popular indoor semantic segmentation benchmark dataset. It contains 10,335 RGB-D images collected across 37 scene categories. A label reduction process was conducted to convert this semantic dataset into a drivable area dataset. First, this study filtered out samples from [32] whose areas contain less than 1% of "floor" or "carpet" labels. Second, the labels of the remaining 7,946 samples were converted into binary maps ("floor" & "floor_mat" =1, all others =0). These samples were then randomly divided into a training set and an evaluation set at a ratio of 4:1. The retrained models [16] and [17] achieved training accuracies (mIoU) of 82.4% and 88.0%, respectively.

For further comparison, vanilla versions of OSTR [13], RedNet, and ESANet were also pre-trained and included in the benchmark as baselines. While models [16] and [17] utilized RGB and depth images as inputs during the tests, [13] and

TABLE IV
QUANTITATIVE RESULTS FROM THE VALIDATION STUDY

| Method | Backbone | MIoU | FPS |
|---|---|---|---|
| Ground Plane Seg | - | 82.7 | 100* |
| **MOSTS** | **MobNetv3** | **85.7** | **93.13 (48.22)** |
| MOSTS (Voronoi) | MobNetv3 | 79.1 | 93.13 (48.22) |
| MOSTS-Res34 | Res34 | 86.2 | 22.77 (18.54) |
| MOSTS-Res50 | Res50 | 86.3 | 17.60 (14.97) |
| OSTR [13] | Res50 | 73.0 | 13.44 (11.85) |
| OSTR [13] (Perlin) | Res50 | 83.5 | 13.44 (11.85) |
| RedNet [16] | Res50 | 71.0 | 11.44 |
| ESANet [17] | Res34 | 80.7 | 21.71 |
| RedNet [16] (DAS) | Res50 | 85.1 | 11.44 |
| ESANet [17] (DAS) | Res34 | 83.0 | 21.71 |

The result of the proposed method is **bolded**. *: FPS was estimated using typical algorithm throughput. (): FPS adjusted for the latency of ground plane segmentation algorithm.

MOSTS used the RGB (masked) and texture reference images as inputs. All image data were resized to 256 × 256. Lastly, all tested neural network models were converted into Float16 IR and deployed to the same SBC to estimate the inference speed.

## VII. RESULTS AND DISCUSSION

The performance metrics used in the validation test included accuracy (measured in mIoU) and inference speed (measured in FPS). As Table IV demonstrates, [13] attained an accuracy of 73% on ISOD. Training [13] with the Perlin-Noise-based collage generator (as opposed to the Voronoi-based) resulted in an accuracy boost of +10.5%, while MOSTS trained with the Voronoi-based collage generator exhibited a decrease in accuracy (-6.6%). This indicates that synthetic training examples generated via the Perlin-Noise-based method better prepare models for real-world scenarios.

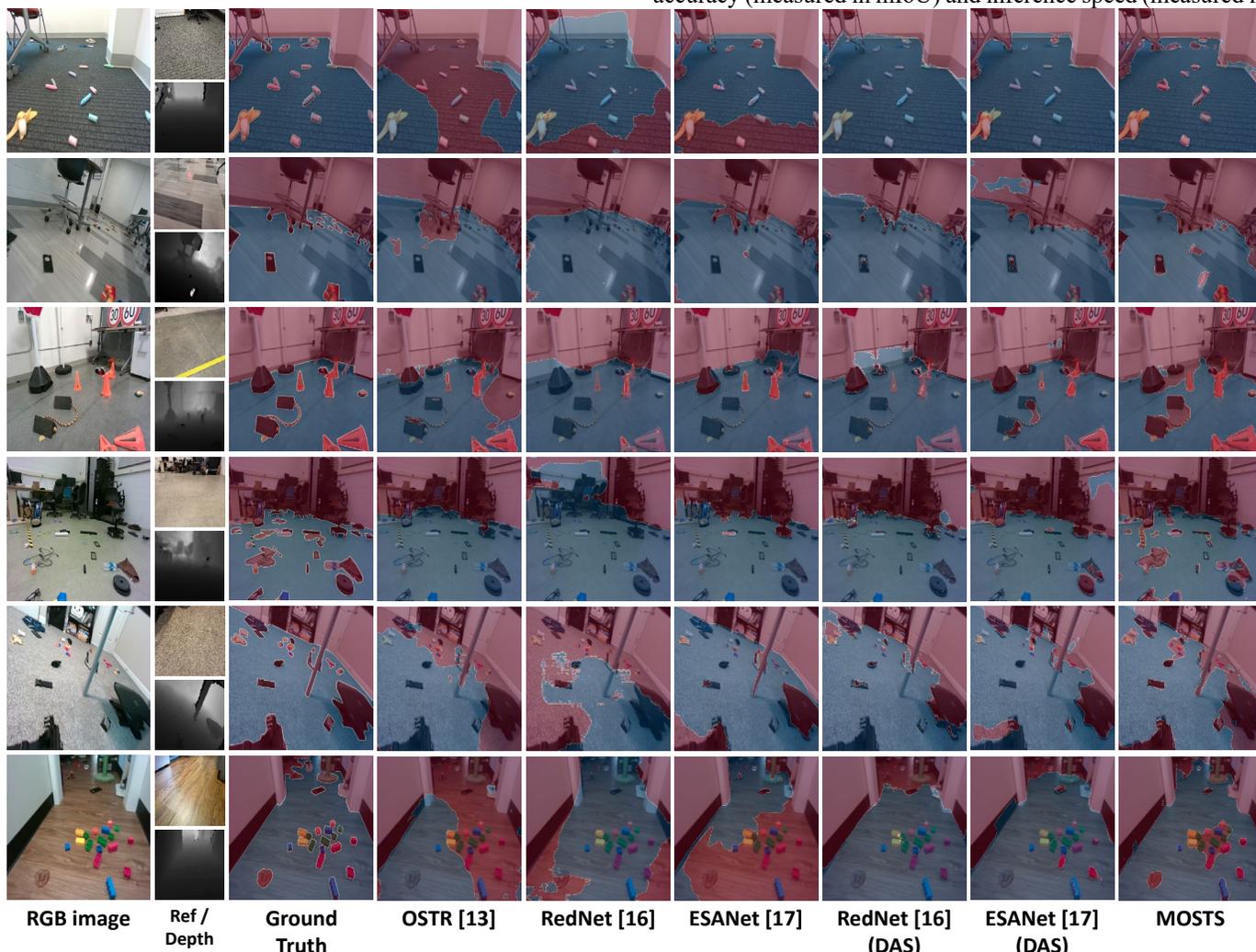

Fig. 12. Qualitative results from the validation test. Drivable areas are highlighted in **blue**, whereas obstacles are highlighted in **red**. Reference images and depth images are shown in smaller size.

The pre-trained vanilla semantic models in [16] and [17] achieved accuracies of 71.0% and 80.7%, respectively. Fine-tuned DAS versions of these models showed improved accuracies, with [16] reaching 85.1% and [17] hitting 83.0%. This suggests [16] possesses greater generalizability, whereas [17] seems to overfit the training data as it has higher training accuracy. By design, [17] is more speed optimized. The inference speed results suggest that [17] is about two times faster than [16]. With an inference speed of 21.71 FPS, [17] will suffice for most indoor robot perception needs. However, MOSTS outperforms [17] in terms of speed by over two-fold while having competitive accuracy, thereby freeing up even more computational resources for subsequent robot control algorithms. The latency of the ground plane segmentation algorithm can be minimized through parallel processing. In that case, MOSTS could attain up to four times faster inference speed (93.13 FPS) than [17], positioning it as a superior choice for indoor DAS.

Further analysis of the performance differences resulting from various backbone selections was conducted, benchmarking MOSTS with ResNet-34 and ResNet-50 as backbones. The surge in inference speed was largely due to the faster backbone. Meanwhile, the accuracies of these ResNet-based MOSTS models are comparable to the proposed one, with MOST-Res50 exhibiting the highest accuracy in this test and an FPS similar to [16].

The validation study also included the accuracy of the ground plane segmentation algorithm. This accuracy was computed by comparing the ground truth with the masked image, where blackened pixels were mapped to 0 and all others to 1. Despite the algorithm's inability to identify small objects ($\leq 5cm$ in height), it achieved a mIoU of 82.7% as the pixel areas of these small objects account for only a small fraction of the entire scene. Thus, the complete omission of these objects does not necessarily result in a significant mIoU reduction. This logic is also applicable to the results of the neural network models. This paper includes some qualitative results from the validation test to gain better insights and further demonstrate the performance differences between the tested methods.

Fig.12 illustrates that ISOD poses significant challenges for indoor semantic models. Both [16] and [17] struggled to identify floors with small objects. The DAS versions demonstrated noticeable improvements, as they had a more general open set for categorizing non-floor objects. While both models managed to capture floor regions even with scene tilting, they largely overlooked smaller objects in the scenes due to the lack of such examples in their training data from SUNRGB-D. In contrast, the proposed method can identify small, low-lying objects irrespective of shape, color, and size. Trained entirely using synthetic images, it doesn't require manually labeled small object training examples like the semantic models. Moreover, as the proposed method first estimates the ground plane using an analytic algorithm, it rarely produces false positive labels on walls and furniture. This safety feature can help reduce the uncertainty inherent in data-driven methods like neural networks.

## VIII. CONCLUSION AND FUTURE WORK

This paper presents a fast, motion-robust RGB-D Drivable Area Segmentation (DAS) framework for indoor robot navigation. This framework uses a ground plane segmentation algorithm to estimate the drivable area, which is further refined using MOSTS, a rapid texture-oriented one-shot segmentation neural network. The MOSTS model was trained using a novel Perlin-noise-based collage synthesis technique. An Indoor Small Objects Dataset (ISOD) was created to evaluate the proposed framework's efficiency. The validation test demonstrated that the proposed method offers competitive accuracy and considerably higher inference speed when compared to two state-of-the-art efficient RGB-D semantic segmentation models. Moreover, the proposed framework is deployable on a low-power Single Board Computer (SBC), capable of delivering real-time performance.

Despite the promising results, future studies are needed to handle the limitations of the proposed method. The framework performs best when a single general floor texture type (e.g., tiled, wooden, carpeted) is present in the scene but suffers from floors featuring mixed texture types.

The potential applications of MOSTS extend beyond refining DAS. For example, Fig.3 shows that MOSTS can effectively detect trash and litter on outdoor terrains like sand and snow. When given an image of well-maintained asphalt, MOSTS can identify cracks on a runway image. It demonstrates that MOSTS can facilitate applications like autonomous cleaning robots and robot runway patrol. Importantly, the same MOSTS model used in the validation test was directly transferred (without any retraining or fine-tuning) to generate the predictions in Fig.3. This illustrates the flexibility of MOSTS, as it can be easily repurposed and adapted to new environments or tasks.